\documentclass[a4paper,runningheads]{llncs}

\RequirePackage{amsmath,amsfonts,amssymb}
\newcommand{\degree}{\ensuremath ^\circ}
\usepackage{bm}
\usepackage{xcolor}
\usepackage{hyperref}
\usepackage{cleveref}
\crefformat{footnote}{#2\footnotemark[#1]#3}

\usepackage{graphicx}
\graphicspath{{}}

\begin{document}

\title{Experimental study of time series forecasting methods for groundwater level prediction}

\titlerunning{Groundwater level forecasting}

\author{Michael Franklin Mbouopda\inst{1}
 \and
 Thomas Guyet\inst{2}
 \\
 Nicolas Labroche\inst{3}
 \and
 Abel Henriot\inst{4}
}

\authorrunning{Mbouopda et al.}

\institute{University Clermont Auvergne, Clermont Auvergne INP, CNRS, Mines Saint-Etienne, LIMOS, Clermont-Ferrand, France
\and
Inria -- Centre Inria de Lyon
\and
Université de Tours
\and
BRGM
}

\maketitle 

\begin{abstract}
Groundwater level prediction is an applied time series forecasting task with important social impacts to optimize water management as well as preventing some natural disasters: for instance, floods or severe droughts. Machine learning methods have been reported in the literature to achieve this task, but they are only focused on the forecast of the groundwater level at a single location. A global forecasting method aims at exploiting the groundwater level time series from a wide range of locations to produce predictions at a single place or at several places at a time. Given the recent success of global forecasting methods in prestigious competitions,
it is meaningful to assess them on groundwater level prediction and see how they are compared to local methods. 
In this work, we created a dataset of 1026 groundwater level time series. Each time series is made of daily measurements of groundwater levels and two exogenous variables, rainfall and evapotranspiration.
This dataset is made available to the communities for reproducibility and further evaluation. 
To identify the best configuration to effectively predict groundwater level for the complete set of time series, we compared different predictors including local and global time series forecasting methods. We assessed the impact of exogenous variables.
Our result analysis shows that the best predictions are obtained by training a global method on past groundwater levels and rainfall data.

\end{abstract}

\section{Introduction}
Groundwaters are water bodies entrapped in soils and rocks, representing storage of significant volumes of water, usually preserved from pollutants. Generally speaking, rainfall would penetrate through soils, rocks and tends to increase the volume of groundwater, whereas drainage by springs or rivers, or water abstraction by pumping would decrease the volume of available groundwater in a given reservoir. Wells and boreholes are the devices capable of monitoring the level of water in the reservoir, and are more generally designated under the term of \textit{piezometers}.
Forecasting groundwater levels contributes to the responsible management of an essential resource for different uses — human consumption (two-thirds of the water supplies for human consumption come from underground sources), irrigation, industrial uses (cooling, washing, etc.) -- but also to watercourse flow management \cite{rahman2020multiscale,rodriguez2014predictive}. 

This article is focused on comparing local and global forecasting methods for predicting the evolution of groundwater levels at single or multiple locations. 
In France, groundwater levels records are stored in a publicly available database (ADES) operated by BRGM\footnote{BRGM: Bureau des Recherches Géologiques et Minières (French geological survey)}.
Data can be requested using public API's through the Hub'Eau portal \url{http://hubeau.eaufrance.fr/}. 
The database stores a network of approximately twenty thousand wells with at least one measure, but for more than a thousand  of those piezometers, several years of continuous data are available, with only a few missing data. 
The prediction horizon was set to about three months (90 days). This corresponds to a forecast of the medium-term evolution of the groundwater levels. In terms of water management, this length of time corresponds to the needs of drought anticipation, usually performed in the late spring after the winter recharge period. In general terms, this timescale corresponds to the vast majority of the needs of applied water management.

The prediction of these levels is a true challenge given the complexity of the hydrological mechanisms at play \cite{Bredy_2020}. Many numerical models have been developed for this purpose. Certain models are based on physical groundwater modelling. Physical models require their parameters to be adjusted to each situation \cite{nayak2006groundwater}. 
They offer an accurate prediction solution but are generally difficult to generalize on a large scale. For more systematic prediction, the use of past groundwater level time series has been viewed for years as an essential tool for water resource planning~\cite{kisi2012forecasting}. 
Many machine learning models have therefore been developed to predict groundwater levels. For instance, Br\'edy et al. \cite{Bredy_2020} put forward a modelling approach for groundwater level forecasting based on two decision-tree-based models, namely Random Forest (RF) and Extreme Gradient Boosting (XGB). Rahman et al.~\cite{rahman2020multiscale} combined wavelet transform with random forest and gradient boosting trees to predict groundwater level at scale in the city of Kumamoto in Japan. Osman et al.~\cite{Ahmedbahaaaldin2021} also evaluated an XGB model but also support vector regression (SVR) and neural networks. 
In these studies, data includes past groundwater level values as well as rainfall and evapotranspiration data. With this information, two natural groundwater inflows and outflows are taken into account. Note that we do not have information about water usage that is difficult to collect from open resources.

Among the many state-of-the-art models, it is difficult to identify the best one for predicting the evolution of groundwater levels. Therefore, it turns out to be relevant to explore various models designed to predict the evolution of time series for the specific data provided and compare them to identify the best model.

In this article, we focus specifically on the following question: ``\textit{Is it preferable to train a model on measurements from a single sensor (local model) or to train a global model to forecast the evolution of piezometric levels?}''. The usual time series forecasting approaches are designed to train models to predict future measurements from a sensor based on past measurements by the same sensor. However, a model trained on each piezometer would require a large volume of past data for each device. Such an approach would rapidly fail as the amount of available data is limited. This is also not the case for the installation of new sensors for example, and more generally it means that the model has low robustness to changes in the functioning of the hydrological system measured. The use of models trained on several piezometers can ensure greater robustness and less training effort for new piezometers. Recently, global forecasting methods (GFM) has shown astonishing results by achieving the best score on distinguished sales prediction competitions \cite{Makridakis2020M4,Makridakis2022M5}, however no prior work on groundwater level forecasting has considered these approaches yet. 
In aquifers, groundwater level fluctuations are not completely independent since water flows from high to low altitudes, along flow paths that depend on the aquifer geometry, porosity (the amount of connected void in the soils and rocks) and thus types of rocks, geometry of the recharge area, or discharge area, unsaturated zone thickness, etc. Inside a given aquifer, it is then expected that groundwater levels at close locations or best, along a same flow path, would have a strong co-linearity, whereas this similarity would fade with increasing distance. For distinct aquifers, it is expected that a similar groundwater level signal could be observed when input signals (rainfall, effective rainfall) are similar or close, with a certain proportion of divergence due to distinct aquifers properties. In simpler words, a proportion of redundant signal can be found along all the groundwater level records, the remaining signal being site specific. 
These reasons make us hypothesize that GFM would be more effective on groundwater level forecasting than local forecasting methods (LFM).
Our exploration of models designed to predict the evolution of groundwater level is therefore structured around the goal of comparing two modelling strategies: local vs global methods.

For each of these types of modelling, we put forward several possible implementations and compared them with respect to the root mean squared scaled error. The rest of this paper is organized as follows: The data used in this study are described in Section~\ref{sec:data-collection}. Section~\ref{sec:gwl-forecasting} presents the formalization of the studied problem and the resolution methods. The conducted experiment and the results are detailed in Section~\ref{sec:experiment}. This work is finally concluded in Section~\ref{sec:conclusion}.

\section{Data collection}\label{sec:data-collection}
\begin{figure}[tbp]
        \centering
        \includegraphics[width=\textwidth]{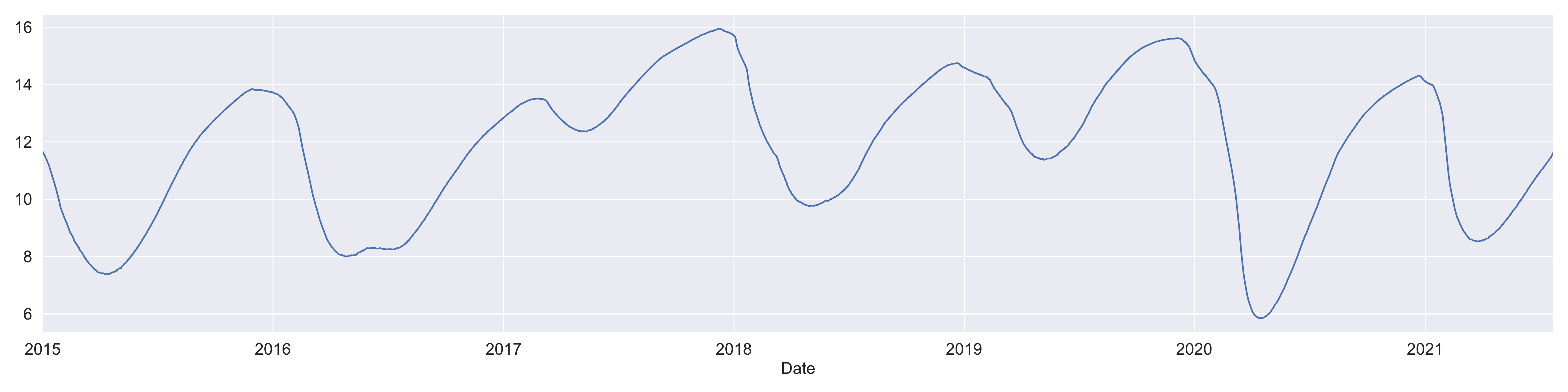}
        \includegraphics[width=\textwidth]{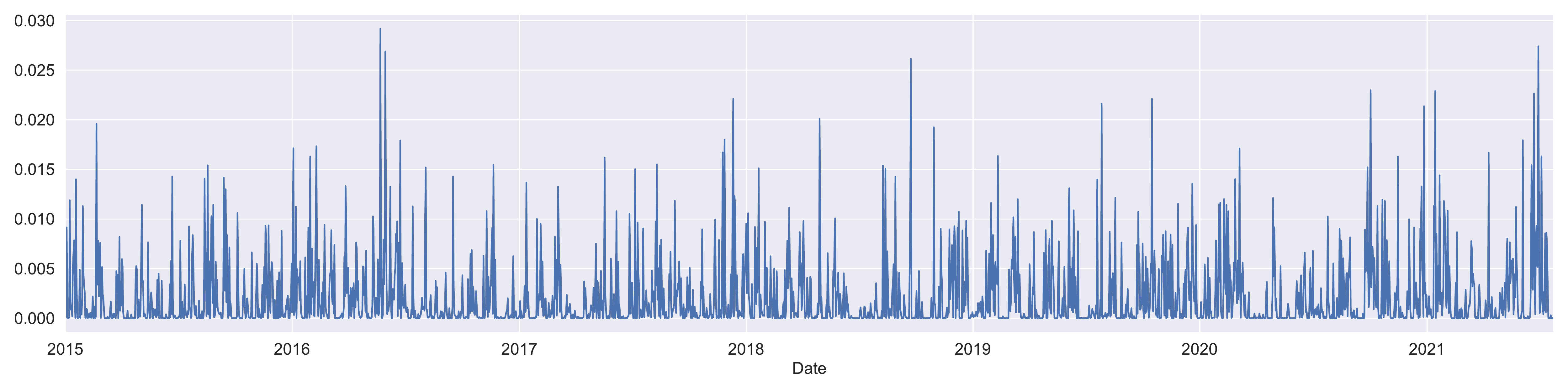}
        \includegraphics[width=\textwidth]{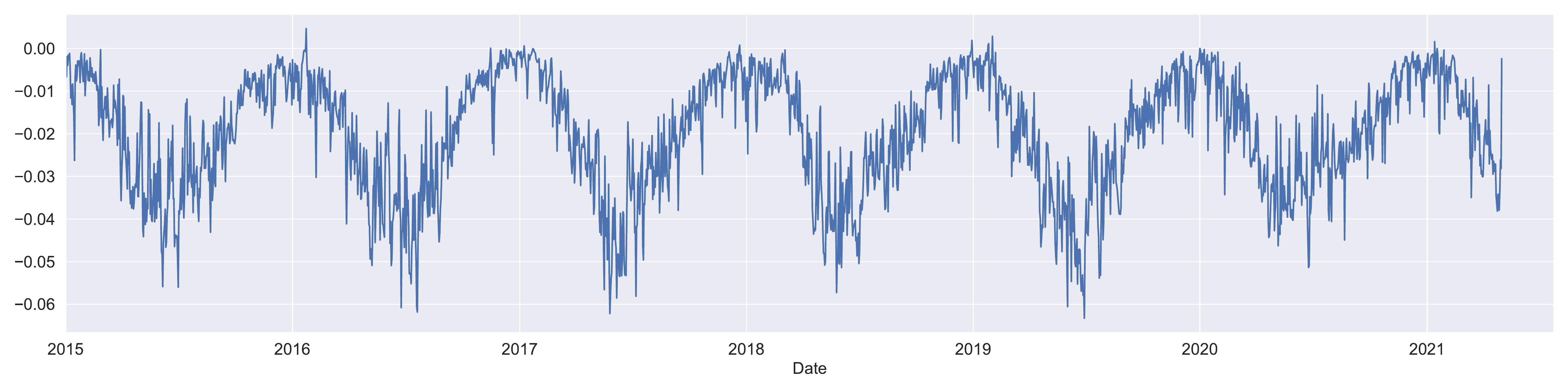}
    \caption{From top to down, daily groundwater level, rainfall and evapotranspiration time series for the piezometer \textsf{BSS000EBLL}.}
    \label{fig:sample-gwl}
\end{figure}

For this study, we compiled a dataset for the period from January 2015 to January 2021 (2,221 days) for a subset of 1,026 piezometers in the French mainland. 
In the following, we explain the selection of the subset of piezometers. 
Each piezometer corresponds to a multivariate time series. 
The multivariate time series has three variables: the groundwater level and two exogenous weather variables: rainfall and evapotranspiration (ETO). 
These latter values have been collected from ERA5 archives using the Climate Data Store (CDS) API~\cite{hersbach2018era5} at the location of the piezometer. This simple approach has been considered accurate enough in the context of this work, even if a more realistic approach had considered the recharge area. 
Nevertheless, considering a weather variable computed at the scale of each recharge area would have led to a significant increase in the complexity of the models.
In addition, weather variables considered at this spatial resolution do not vary significantly with small distances, then the value at the location of the piezometer seems to be an accurate approximation. 
CDS provides daily weather variable measurements with a spatial resolution of $0.25\degree$ for rain and $0.1\degree$ for ETO. 
Hence, the three variables (groundwater level, rainfall and evapotranspiration) are available daily for each piezometer. 
Figure~\ref{fig:sample-gwl} shows the data for the piezometer \textsf{BSS000EBLL} installed in the city of \textit{Senlis-le-Sec}.
It illustrates a time series that is regular: the range of values remains between $7m$ and $16m$ all over the period, and we can identify a yearly seasonality. 
This is not the case for all the piezometers. 
Some of them do not have such seasonality and exhibit brutal changes at some dates.

In addition to groundwater level, precipitation and evapotranspiration, some past works also included sea levels, reservoir levels, and some hydrological, geological, and physiographical factors. However, we used only groundwater level, precipitation and evapotranspiration in this study as these are the three most used inputs for groundwater level forecasting~\cite{tao2022groundwater}.

The selection of the subset of piezometers has been done according to the number of missing values. 
The CDS data have no missing data, but piezometers may be disconnected temporarily from the network and measurements may be missed over long periods. 
We decided to select piezometers with less than 50 days of missing data (over the 2,221 days). 
This value is low and allows selecting a sufficient number of piezometers. In addition, the selected piezometers are spatially spread all over the French mainland. 
As a preprocessing, the missing values were attributed by linear interpolation. Thus, in the sequel, time series are considered without missing values. 

Note that we did not apply any normalization of the time series values. They are provided as it to the forecasting methods.

\section{Groundwater level forecasting}\label{sec:gwl-forecasting}
The groundwater levels database is a set of pairs $\mathcal{Y}=\langle {Y}^k, \bm{Z}^k \rangle$ where ${Y}^k: y^k_{1\dots t}$ is a univariate time series such that $y^k_i\in \mathbb{R}$ for any $i$, and ${Z}^k: \bm{z}^k_{1..\infty}$ is a multivariate time series, so-called exogenous time series data (${Z}: \bm{z}^k_i\in \mathbb{R}^m$, where $m$ is the number of exogenous time series), known as far as date $t$ but also beyond. Time series ${Y}^k$ is the groundwater level series for piezometer $k$, while the exogenous data series corresponds to rainfall and evapotranspiration associated to this piezometer (thus $m=2$ in this study). 

The prediction of a time series ${Y}^k$ for forecast horizon $h$ consists in estimating $y^k_{t+1 \dots t+h}$. For this time series prediction task, the conventional data analysis approach is to construct an autoregressive model. Such a method constructs a function for the prediction of the value at date $t_0$ from the $r$ latest observations of ${Y}$ and $\bm{Z}$. Such a function for the prediction of the next value in the series is denoted $\varphi : \mathbb{R}^{r\times(m+1)} \mapsto \mathbb{R}$. $\varphi$ can thus be seen as a regressor: it predicts a real value from the input characteristics. 

Classical autoregressive models assume that $\varphi$ is a linear regression (AR model), but $\varphi$ can be modelled by any trainable regression function. We denote such methods as a \textsf{generalized} autoregressive model.
Notice that the classical AR model is a special case of the generalized autoregressive model which assures a linear relationship between the variables.

To produce a forecast for horizon $h$, the prediction function is recursively applied $h$ times. We note that it is assumed that the exogenous time series are known in the future. This problem is therefore different from that of forecasting a multivariate time series for which the prediction function outputs the value of the target series together with the  values of the exogenous series. In our problem, we hypothesize that rainfall and evapotranspiration were similar from one year to the next. Consequently, forecasts can be obtained in the future by taking the mean daily values of past years. The approximation error made by this hypothesis is preferable to that of the cumulative errors of a medium-term multivariate recursive prediction.

There exists many methods for constructing the $\varphi$ function in the state-of-the-art and one goal of this work is to compare the most classical on groundwater level forecasting. The second goal is to decide whether it is better to use a local or a global method.

\subsection{Local versus global time series forecasting}
Given a set of time series, we denote by ``local'' the approaches that consider each time series as an independent dataset and build an individual model to forecast the future values of each time series. 
Local time series forecasting has been the de facto approach for time series forecasting. 
However, nowadays, many companies are collecting a large set of time series from similar sources routinely, and training a single model for each individual series is time-consuming, costly, and difficult to maintain; furthermore, possible relationships between these time series are not taken into account by local approaches. 
For these reasons, a new forecasting paradigm called global forecasting has emerged. A Global Forecasting Model (GFM)~\cite{Januschowski2020} is trained on a set of time series and is then used to forecast future values of each time series. In other words, a GFM learns to predict future values given historic values regardless of the sources of  data. Global forecasting is obtaining astonishing results today, by winning prestigious competitions such as the M4 and M5 challenges~\cite{Makridakis2020M4,Makridakis2022M5} and competitions held recently on the Kaggle platform~\cite{Bojer2021}. Former works on global time series forecasting claimed that time series need to be related. 
However, Montero-Manso and  Hyndman~\cite{Montero-Manso2021} shown theoretically that whatever the heterogeneity of time series, there always exists a GFM that is as good as, or even better than any collection of local models. This result has been further supported by an experimental study~\cite{Hewamalage2022}. Since global method has shown good results, especially in forecasting time series of sales, we are wondering if this is also the case for forecasting time series of groundwater level. 

\subsection{Considered methods}
We explored four versions of the generalized autoregressive forecasting method (see above) as well as recent state-of-the-art neural network methods in the field. More specifically, we considered: 
\begin{itemize}
    \item Linear regression. This is the standard method for regression problems. This method is equivalent to using an AR model of order $r$. 
    \item SVR~\cite{awad2015support}. Support vector regression (SVR) models address high-dimensional problems and propose non-linear models. We also used Gaussian kernels. After a few experiments, we set parameter $C=100$. 
    \item Random Forest~\cite{breiman2001random}. This is a bagging method. We opted for a forest of 100 trees. 
    \item Extreme Gradient Boosting~\cite{chen2016xgboost}. This is a tree-based method which uses optimization techniques to improve the calculation efficiency of machine learning. It can thus process large datasets. 
    \item \texttt{DeepAR}~\cite{Salinas2017DeepAR}. This is a probabilistic forecasting method based on autoregressive recurrent neural networks.
    \item \texttt{Prophet}~\cite{Taylor2017Prophet}. A modular regression model with parameters that are intuitive for analysts and that can be easily adjusted using domain knowledge. \texttt{Prophet} is a decomposable model with three components which are trending to model non-periodic changes in the values of time series, seasonality to model periodic changes (weekly, yearly, etc.) and holidays to represent the effect of holidays on the time series values.
    \item \texttt{NeuralProphet}~\cite{Triebe2021NeuralProphet}. This is a neural network-based implementation of the \texttt{Prophet} model with some enhancements. In addition to the trend, seasonality and holidays components, \texttt{NeuralProphet} considers three additional regression components: auto-regression effect on past observations, regression effect of exogenous variables, and regression effect of lagged observations of exogenous variables. By using neural networks to learn each component, activation functions can be used to learn non-linear behaviours in the time series.
\end{itemize}

\section{Experimental settings}\label{sec:experiment}
This section presents our experimental settings. 
We start by presenting in detail the experiments and then give the comparison metrics.

\subsection{Setup}
We organized our experiments by grouping the considered forecasting models in three categories and by following a naming convention to make things easier to follow. 
The three categories are generalized autoregressive models, \texttt{DeepAR}-based models, and \texttt{Prophet}-based models.

\begin{description}
    \item[Generalized autoregressive models] which contains traditional models not necessarily specifically designed for forecasting: eXtreme Gradient Boosting (XGB), Linear regression (LM), Random Forest (RF), and Support Vector Machine (SVR). These regression models consider a vector as input and output a real value. Therefore, we create a dataset by sliding a window of length $r+h$ over our time series. Each position of the sliding window is a sample in our dataset with the first $r$ values (the history) being the input and the last $h$ (the horizon or the forecast) being the output. 
    In the experiments, we fixed the horizon $h$ to $93$ (three months), and did not put any constraint on $r$, however $r$ is generally required to be greater than $h$ in order to expect good forecasting. We tested values from $40$ to $140$ and found that the best value is $100$ or $110$ in average -- this result is in accordance with the results from \cite{lara-benitez2021} which suggests that a history length $1.25 \times h$ leads to the best forecasting. 
    
    \item[DeepAR-based models] composed of two models as \texttt{DeepAR}~\cite{Salinas2017DeepAR} is a global forecasting method. \texttt{DeepAR-L} is the \texttt{DeepAR} model trained locally, meaning that for each individual time series, a new instance of the model is trained on this time series to forecast future values of this individual time series only. On the contrary, \texttt{DeepAR-G} is a \texttt{DeepAR} model trained once on every time series to predict future values of all the time series simultaneously.

    \item[Prophet-based models] made of \texttt{Prophet}~\cite{Taylor2017Prophet} and \texttt{NeuralProphet}~\cite{Triebe2021NeuralProphet}. NeuralProphet is by default a local forecasting model, but can be configured to run as a global method. Therefore, we considered \texttt{NeuralProphet-L} (for the local version of NeuralProphet) and \texttt{NeuralProphet-G} (for the global version). Prophet can only be used as a local forecasting method.
\end{description}

\texttt{DeepAR}, \texttt{Prophet}, and \texttt{NeuralProphet} are specifically designed for time series forecasting: they take as input a time series and predict the user-defined number of values in the future (93 in our case). Unlike the generalized autoregressive models which use a fixed-length history, \texttt{DeepAR}, \texttt{Prophet}, and \texttt{NeuralProphet} learn from the whole past observations of the time series to make predictions.

For each forecasting method, we considered two configurations and defined a naming convention to make things easier to understand:
\begin{itemize}
    \item the first configuration uses only historic groundwater levels in order to forecast the future levels: exogenous data are not used. This configuration is named using the forecaster name. For instance, this configuration is named \texttt{NeuralProphet-L} and \texttt{XGB} when the models used are \texttt{NeuralProphet-L} and XGBoost respectively.
    
    \item the second configuration does not use only historic groundwater levels to predict the future, but also historic rain and/or evapotranspiration (ETO) data as the dynamic of their corresponding phenomenons could have a significant impact on groundwater level. This configuration is named as \texttt{[forecaster]+rain}, \texttt{[forecaster]+eto} and \texttt{[forecaster]+rain+eto} depending on the exogenous variables used: respectively, rain data, evapotranspiration data, or both. For instance, the configuration that uses \texttt{Random Forest}~(\texttt{RF}) to forecast the piezometric level using historic groundwater levels and rain data is named \texttt{RF+rain}. 
    These exogenous variables are times series that made the $\bm{Z}$ variable.  
\end{itemize}

The experiment source code is written in the Python programming language using open source libraries. For \texttt{DeepAR}, we use the implementation provided in the GluonTS library \cite{gluonts_jmlr}; for \texttt{Prophet} and \texttt{NeuralProphet} we use the implementation provided by the authors; we use the official implementation of the eXtreme Gradient Boosting model\footnote{Source code and supplementary material: \url{https://github.com/dmlc/xgboost}}; and for the remaining models we use Scikit-learn~\cite{scikit-learn}. The source code is available as supplementary material and is available on a public repository \footnote{\label{note:github}Source code: \url{https://github.com/frankl1/piezoforecast}}.

\subsection{Comparison metrics}
The methods are compared with respect to their ability to predict the next three months of daily values. More specifically, we analysed the root mean squared scaled error (RMSSE) of the methods over the period from the 15th of October 2021 to the 15th of January 2022, corresponding to a horizon of 93 days (roughly three months). This period is not used during the training phase of the methods. Indeed, predictions may be easier or more difficult to perform depending on the period of the year. For instance, during dry periods, groundwater is less affected by rainfall, for example. However, we focused on these three months horizon as it  corresponds to the needs of drought anticipation and is also used in the literature~\cite{rahman2020multiscale}. For a time series of length $n$ and a forecasting horizon $h$, the RMSSE is defined as follows:

\begin{equation}\label{eq:rmsse}
    RMSSE=\sqrt{ \frac{\frac{1}{h}\sum_{t=n+1}^{n+h}\left(y_t-\hat{y_t}\right)^2}{\frac{1}{n-1}\sum_{t=2}^{n}\left(y_t-y_{t-1}\right)^2}}
\end{equation}

In addition to the RMSSE, we assess the significance of the difference between the methods. This evaluation is summarized using critical difference diagrams~\cite{demsar2006statistical} by rejecting the null hypothesis using the Friedman test followed by a pairwise post-hoc analysis as recommended by Benavoli et al.~\cite{benavoli2016}. The critical difference diagrams are drawn using the source code made public by Fawaz et al.~\cite{fawaz@2019}. Elsewhere, we also use box plots to summarize our results. Detailed results are available in the accompanying repository of this paper\cref{note:github}.

\section{Results}
\subsection{Generalized autoregressive models results}
\begin{figure}[tbh]
    \centering
    \includegraphics[width=.8\textwidth]{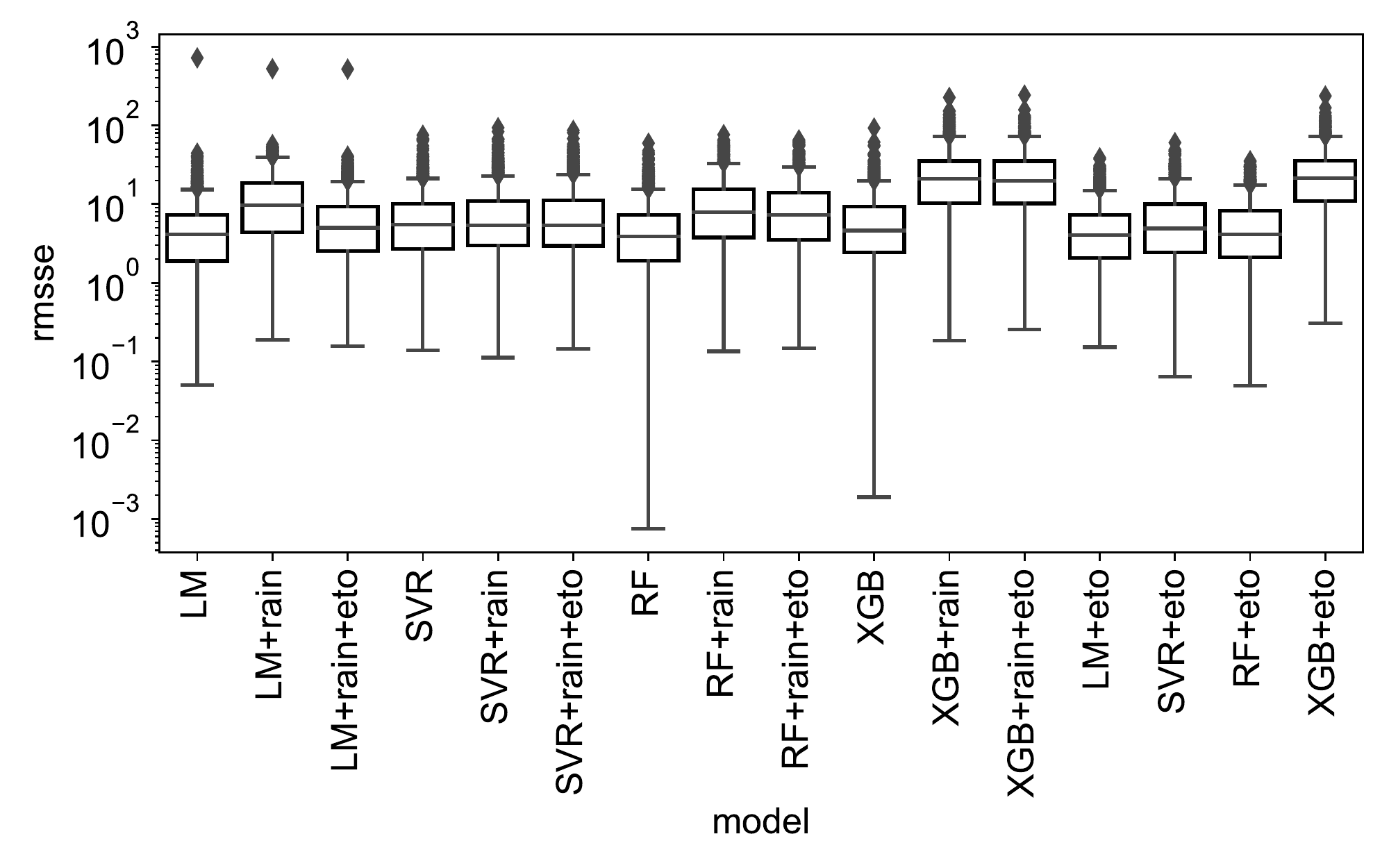}
    \caption{RMSSE of the generalized autoregressive models.}
    \label{fig:boxplot-legacy}
\end{figure}

\begin{figure}[tbh]
    \centering
    \includegraphics[width=\textwidth]{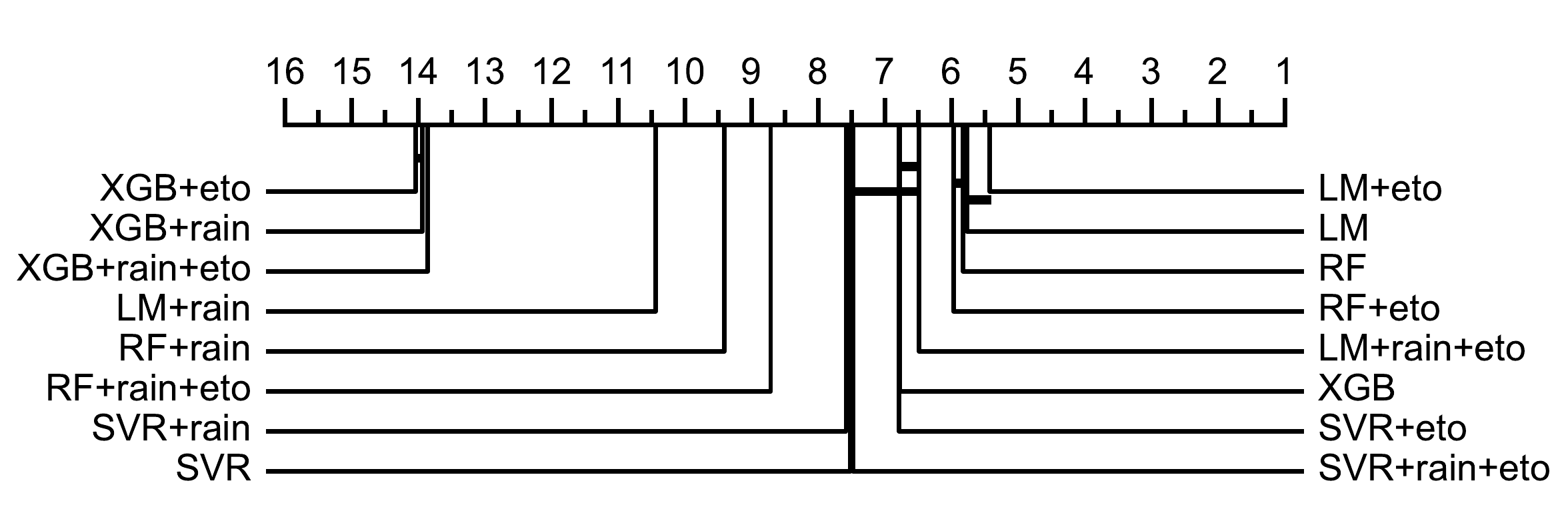}
    \caption{Critical difference diagram of the generalized autoregressive models.}
    \label{fig:cd-legacy}
\end{figure}

The performance achieved by the generalized autoregressive models are summarized on Figure~\ref{fig:boxplot-legacy} and the significance of the difference between each pair of models is depicted on Figure~\ref{fig:cd-legacy}.

It can be observed that the best performing configurations use either only evapotranspiration as exogenous data, or no exogenous data at all. These configurations are made of linear models (see \texttt{LM} and \texttt{LM+eto}) and random forests (see \texttt{RF} and \texttt{RF+eto}). Although the difference is not significant, \texttt{LM+eto} performs better than \texttt{LM}. On the contrary, \texttt{RF} performs better than \texttt{RF+eto}. When there are no exogenous data, \texttt{LM} is the best model.

\texttt{SVR}, \texttt{SVR+rain} and \texttt{SVR+rain+eto} have very similar RMSSE in average, meaning that the impact of exogenous data is not significant for SVR-based configurations. On the contrary, exogenous data have a significant negative impact on tree-based configurations (\texttt{RF} and \texttt{XGB}) since XGB+rain and \texttt{XGB+rain+eto}, \texttt{RF+rain}, and \texttt{RF+rain+eto} achieve the highest RMSSE while \texttt{XGB} and \texttt{RF} are among the top four best configurations.

It is observed that the left part of Figure~\ref{fig:cd-legacy} is mainly made of configurations that use exogenous data, while the right part is mainly composed of a mix of configurations that use or do not use exogenous data. Using exogenous data decreases the performance of LM, XGB (\texttt{XGB+eto}, \texttt{XGB+rain}, and \texttt{XGB+rain+eto} are the worst configurations), \texttt{RF}, but increases performances of \texttt{SVR} (see \texttt{SVR+eto} and \texttt{SVR+rain+eto}). Therefore, using exogenous data makes the forecast better or worse, depending on the used forecaster and its ability to capture complex relationships.

\subsection{DeepAR-based models results}
\begin{figure}[tbh]
    \centering
    \includegraphics[width=.7\textwidth]{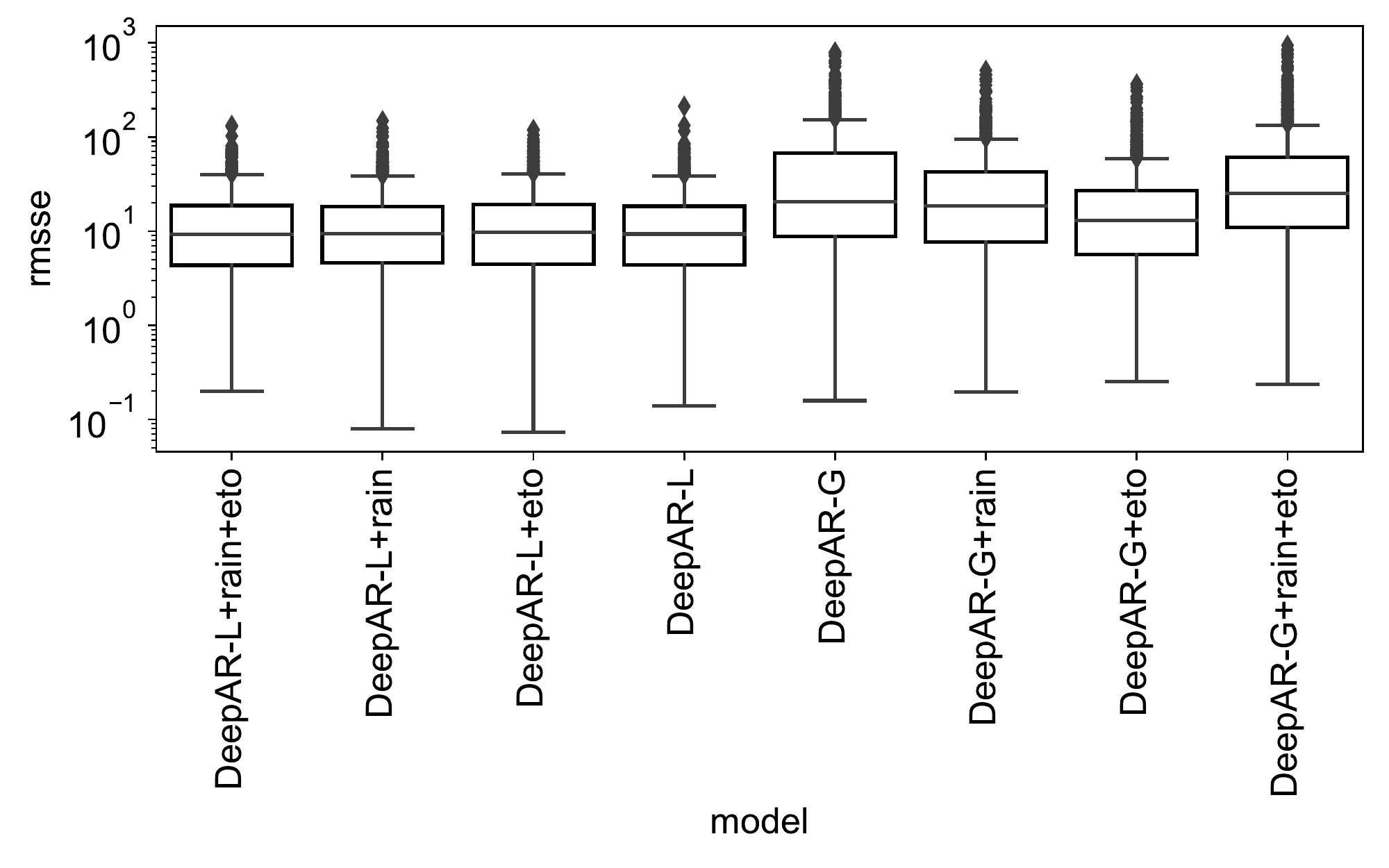}
    \caption{Box plots summary of \texttt{DeepAR} models results.}
    \label{fig:boxplot-deepar}
\end{figure}

\begin{figure}[tbh]
    \centering
    \includegraphics[width=\textwidth]{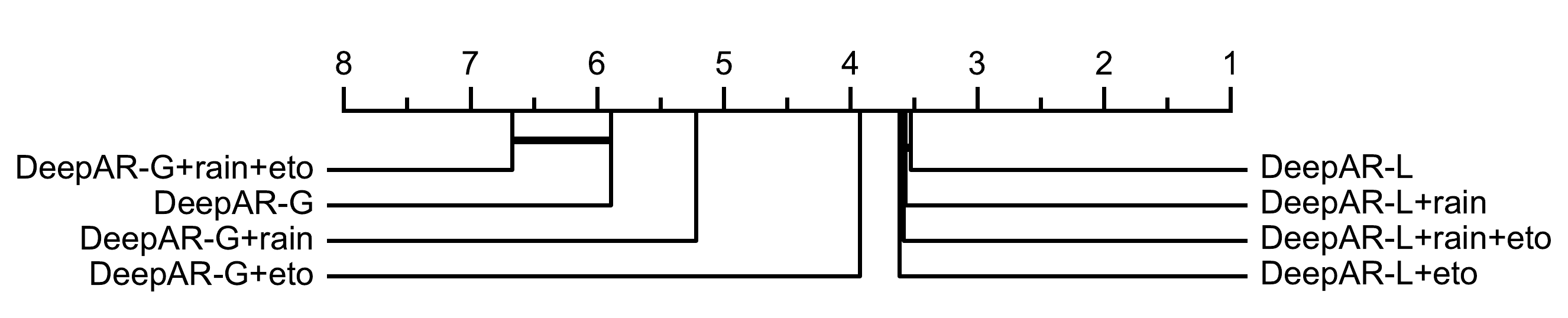}
    \caption{Critical difference diagram of \texttt{DeepAR}-based models.}
    \label{fig:cd-deepar}
\end{figure}

The results obtained using \texttt{DeepAR} as local method as well as a global method are shown in Figure~\ref{fig:boxplot-deepar} and Figure~\ref{fig:cd-deepar}. 

\texttt{DeepAR-L} produces predictions that are significantly better than predictions of \texttt{DeepAR-G}, and this regardless of using exogenous data or not. This suggests that a local training strategy for \texttt{DeepAR} is better than a global one for groundwater level prediction. 

Although \texttt{DeepAR} is naturally designed to take advantage of exogenous data when available, it can be observed that rain and/or evapotranspiration have negative, yet insignificant impact on \texttt{DeepAR-L}'s predictions. On the contrary, exogenous data significantly improve \texttt{DeepAR-G}'s predictions, except when rain and evapotranspiration are used simultaneously. 

\subsection{Prophet-based models results}
\begin{figure}[tbh]
    \centering
    \includegraphics[width=.7\textwidth]{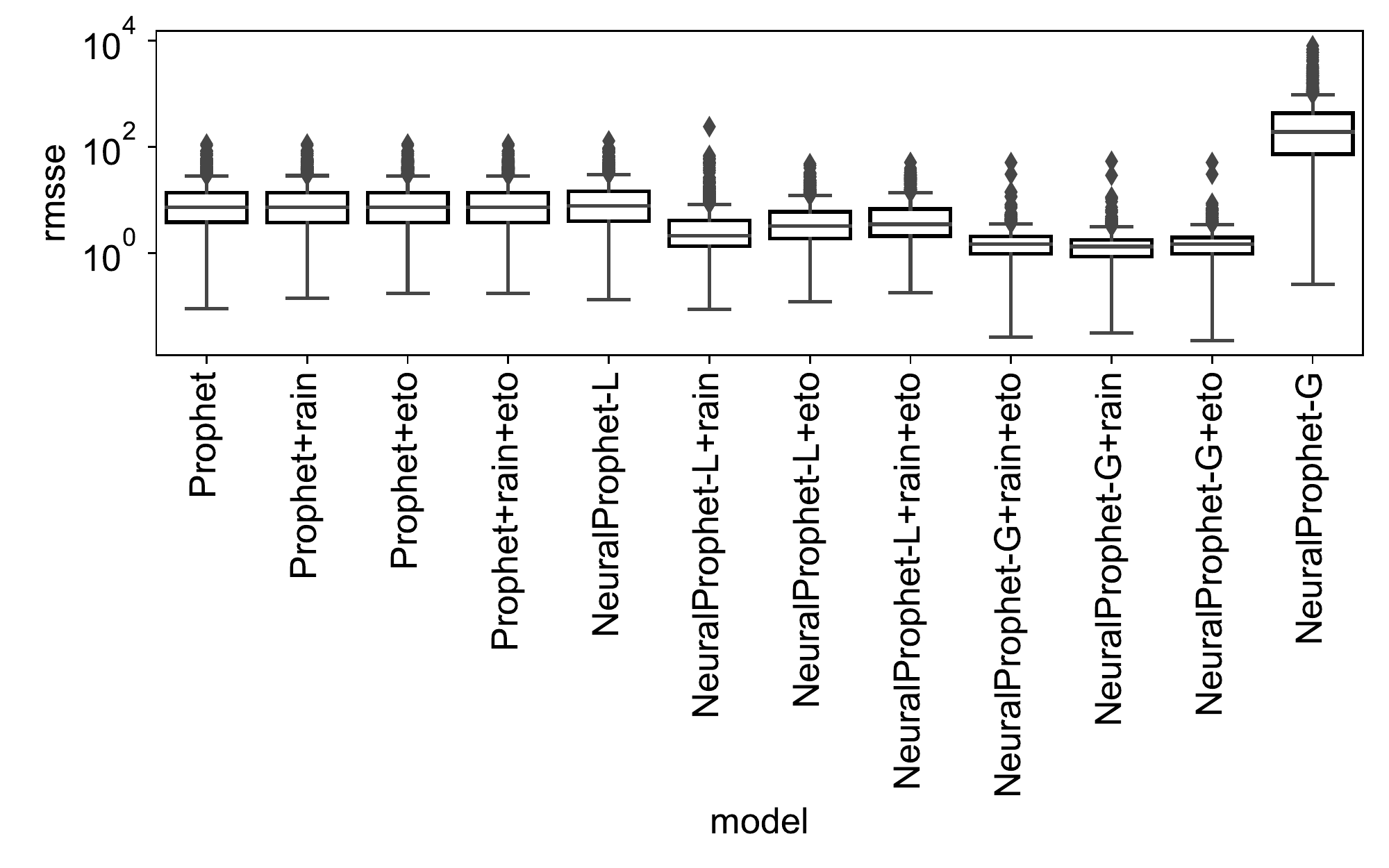}
    \caption{Box plots summary of \texttt{Prophet} and \texttt{NeuralProphet} results.}
    \label{fig:boxplot-prophet}
\end{figure}

\begin{figure}[tbh]
    \centering
    \includegraphics[width=\textwidth]{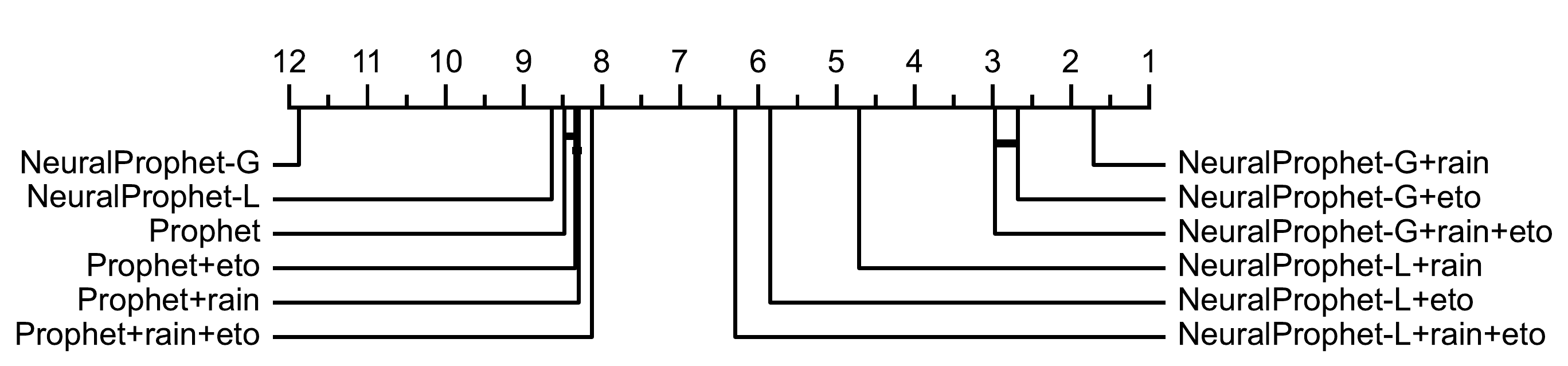}
    \caption{Critical difference diagram of \texttt{Prophet}-based models.}
    \label{fig:cd-prophet}
\end{figure}

The comparison of \texttt{Prophet} and \texttt{NeuralProphet}-based configurations are displayed in Figure~\ref{fig:boxplot-prophet} and Figure~\ref{fig:cd-prophet}.

Unexpectedly, \texttt{NeuralProphet-G} produces forecasts that are worse than those obtained by any other \texttt{NeuralProphet} or \texttt{Prophet}-based configurations. However, using exogenous data (rain and/or evapotranspiration) makes \texttt{Neural\-Prophet-G} the best forecaster, particularly \texttt{NeuralProphet-G+rain}. This result suggests that training a global \texttt{NeuralProphet} model for groundwater forecasting achieves more reliable predictions when exogenous data are used.

Unlike \texttt{DeepAR}, \texttt{NeuralPhophet} works better when executed globally than locally, in particular when exogenous data are used. Furthermore, using exogenous data has a positive impact on \texttt{Prophet}, \texttt{NeuralProphet-L}, and \texttt{NeuralProphet-G}. This impact is not significant for \texttt{Prophet}, but it is for \texttt{NeuralProphet}.

In the absence of exogenous data, \texttt{Prophet} significantly outperforms \texttt{Neural\-Prophet} at predicting the future values of groundwater levels.

\subsection{Comparing the three groups of models}
\begin{figure}[tbh]
    \centering
    \includegraphics[width=\textwidth]{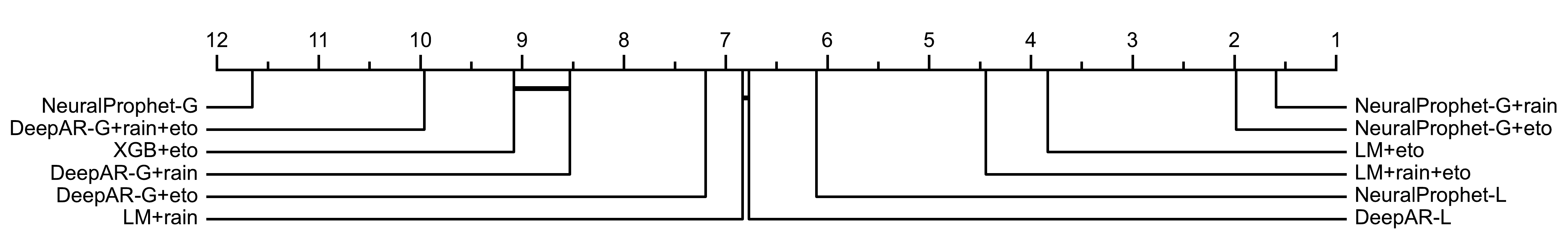}
    \caption{Critical difference diagram comparing the representatives of groups.}
    \label{fig:cd-best}
\end{figure}

The previous section discussed the performances of the three groups of models separately. It compares them to each other. For the sake of readability, we selected some ``representative'' methods from each group. However, an exhaustive comparison is available as supplementary material\cref{note:github}. 
The representative models for each category are the two best and the two worst configurations. Also, we make sure that predictions of the selected methods are significantly different from each other regarding the critical difference diagram of the corresponding group. The comparison of the representatives methods is depicted in Figure~\ref{fig:cd-best}.

The linear model (\texttt{LM}) and \texttt{NeuralProphet-G} respectively produce the most accurate and the least accurate forecasts when exogenous data are absent. The configuration \texttt{NeuralProphet-G} significantly outperforms the others when exogenous data are used, with \texttt{NeuralProphet-G+rain} being the best of all. The configuration \texttt{NeuralProphet-G+rain} is significantly better than the configuration \texttt{NeuralProphet-G+eto}, suggesting that precipitations impact groundwater level more than evapotranspiration does. However, as \texttt{LM+eto} outperforms \texttt{LM+rain}, the impact of exogenous data depends on the capabilities of the model to exploit these additional sources of knowledge.

Except for \texttt{NeuralProphet-G}, \texttt{NeuralProphet}-based configurations are better than \texttt{DeepAR}-based configurations at forecasting groundwater levels, especially when exogenous data are available and the training is done on every time series (i.e. globally).

\texttt{DeepAR-L} and \texttt{LM+rain} are not significantly different from each other in terms of RMSSE. The same observation is true for \texttt{XGB+rain+eto} and \texttt{DeepAR-G+rain}. This result suggests that no long-term dependency is lost by using a history of length $100$. In other words, the time dependency in groundwater level data is not longer than $100$ time steps when focused on the prediction at a horizon of 90 time steps.

\begin{figure}[tbh]
    \centering
    \includegraphics[width=\textwidth]{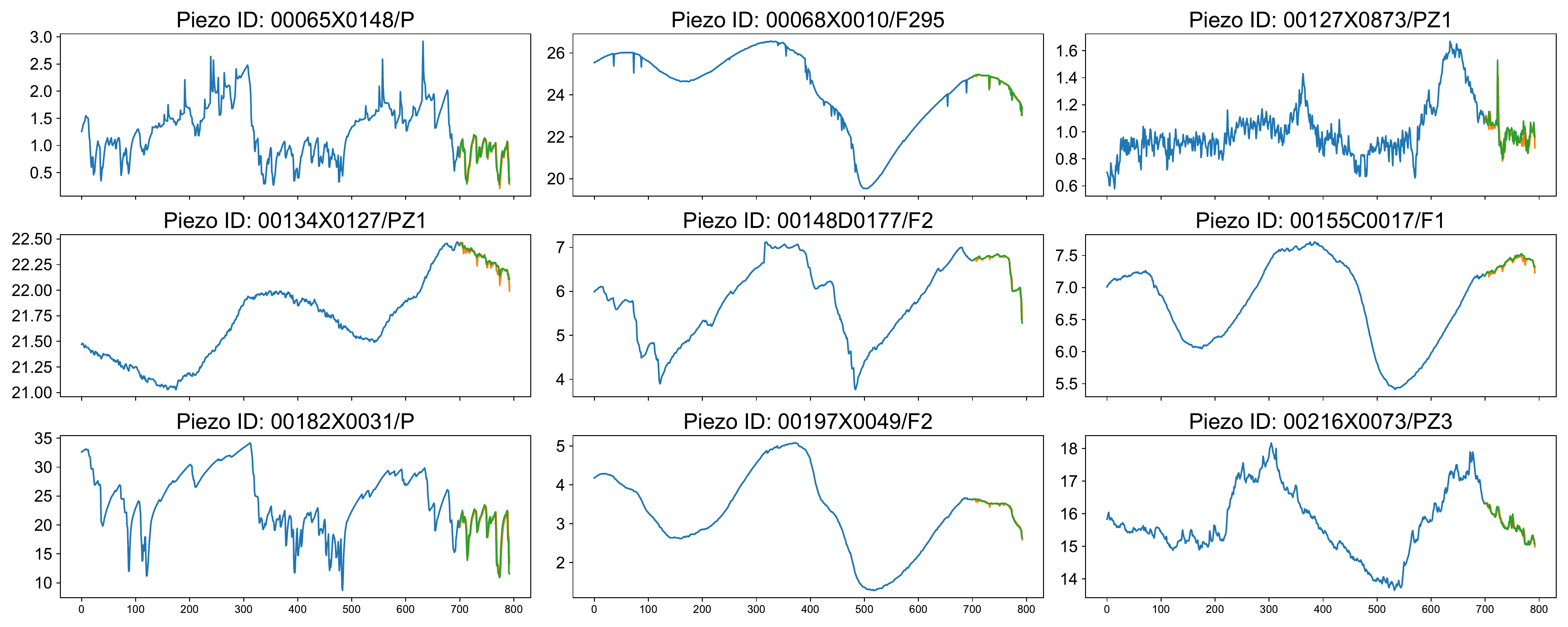}
    \caption{Some forecasts obtained using the configuration \texttt{NeuralProphet-G+rain}.}
    \label{fig:some-forecasting}
\end{figure}

Figure~\ref{fig:some-forecasting} shows predictions obtained with \texttt{NeuralProphet-G+rain} for some piezometers. We observe that forecasts are very close to ground truth groundwater levels.

\subsection{Discussion}
We have compared a set of state-of-the-art forecasting methods for groundwater level prediction in order to come up with the best method. We have considered local forecasting methods as well as global methods and we evaluated the impact of using exogenous data. 
Our finding is that selection of the best model is highly dataset dependent. When no exogenous data are considered, LM models perform better than more complex models. Nevertheless, these models do not appear as the more useful, as they cannot efficiently afford any prediction given a set of weather prediction (drought/flood). They can still be used when no rain or evapotranspiration data are available at a given location.
When considering exogenous data, \texttt{NeuralProphet-G+rain} has been found to be the best model among the set we tested. 
In this particular case, the global model tends to outperform local models. 
It catches useful information in the redundancy found in the different piezometers. Our intuition is that complex processes are at play on groundwater levels (trends, seasonality). They have to be taken into account in forecasting models, for instance with \texttt{NeuralProphet} models. 
In addition, this type of model is rather simple to implement and could be easily extended or enhanced for a larger dataset.
Nonetheless, this training strategy does not work with all the forecaster, e.g. with \texttt{DeepAR}.

We did not consider a couple of aspects of this study and that could improve groundwater level predictions. 

The first limitation is related to the choice of the methods compared and the datasets. On the one hand, we acknowledge that there exist many more forecasting methods in the literature in addition to what we considered in this study. But as we wanted to evaluate the performance of global methods trained on a large collection of groundwater level time series, we selected the most effective forecasting methods in the literature. On the other hand, the dataset used in this study is only representative of the French groundwater levels -- there is no guarantee that our results generalize to a different country. Keeping this in mind, we made our source public to allow our work to be extended easily. Nevertheless, as global methods achieved the best performance, we believe that they will generalize at least as well as any local method.

The second limitation is that we did not search for the optimal hyperparameter values for each method. Instead, we used the default parameters of the implementation used. 
Although we could have fixed this limitation using techniques such as a grid search and Bayesian optimization. We chose a simple approach with the default values because performing hyperparameter tuning for such a large number of datasets (1,026 multiplied by the number of methods) would require a lot of time and resources. However, knowing how methods are compared to each other without hyperparameters tuning could be seen as a measure of the effort required by methods to be effective; in other words, how many are methods ``plug-and-playable''.

Finally, global forecasting methods we evaluated in this work have been specifically designed for global forecasting. We have not assessed the capabilities of generalized autoregressive methods when trained globally. Although assessed methods were enough to evaluate our hypothesis about the effectiveness of GFM for groundwater level forecasting, comparing generalized autoregressive methods when trained globally with specifically designed GFM is worthy. 

\section{Conclusion}\label{sec:conclusion}
In this work, we performed an experimental study consisting of evaluating several methods for predicting the evolution of groundwater levels for 1,026 different piezometers. We considered local and global forecasting methods, and we compared them regarding the root mean squared scaled error metrics. The conducted experiments confirmed our hypothesis that global forecasting methods could be more effective than local ones for groundwater level prediction. In particular, global \texttt{NeuralProphet} achieved the best predictions when using past groundwater levels and rainfall as input. However, when only past groundwater levels are available (no rainfall and no evapotranspiration) local methods and particularly the linear regression model achieve the best predictions. As mentioned in the discussion section, this work could be improved. In particular, we are planning to assess the performance of generalized autoregressive methods (XGB, RF, LM, SVR, etc.) when they are trained globally and comparing them to specifically designed global methods. 

\bibliographystyle{splncs04}
\bibliography{biblio}
\end{document}